\documentclass{INTERSPEECH2023}

\usepackage{makecell}
\usepackage{url}
\emergencystretch=\maxdimen
\interspeechcameraready

\title{N-gram Boosting: Improving Contextual Biasing with Normalized N-gram Targets}
\name{Wang Yau Li, Shreekantha Nadig, Karol Chang, Zafarullah Mahmood, Riqiang Wang, Simon Vandieken, Jonas Robertson, Fred Mailhot}
\address{
  Dialpad  Inc., Canada
  }
\email{\{wangyau.li,shreekantha.nadig,karol.chang,zmahmood,riqiang.wang,
svandieken,jonas,fred.mailhot\}@dialpad.com}

\begin{document}

\maketitle
 
\begin{abstract}
Accurate transcription of proper names and technical terms is particularly important in speech-to-text applications for business conversations. These words, which are essential to understanding the conversation, are often rare and therefore likely to be under-represented in text and audio training data, creating a significant challenge in this domain. We present a two-step keyword boosting mechanism that successfully works on normalized unigrams and n-grams rather than just single tokens, which eliminates missing hits issues with boosting raw targets. In addition, we show how adjusting the boosting weight logic avoids over-boosting multi-token keywords. This improves our keyword recognition rate by 26\% relative on our proprietary in-domain dataset and 2\% on \mbox{LibriSpeech}. This method is particularly useful on targets that involve non-alphabetic characters or have non-standard pronunciations.
\end{abstract}

\section{Introduction}

Recognition of conversationally relevant keywords such as names and technical terms (e.g. ``X-mAbs'') is difficult due to their rarity, but it is particularly important in the context of business communication. Keywords usually highlight the key topics of the conversation and are thus essential to understanding the conversation. Incorporating contextual information is critical to improving the performance of automatic speech recognition (ASR) systems \cite{aleksic15_interspeech}, which allows the system to favour certain keywords that would be more probable to appear in a given context. This information can be included statically or dynamically depending on the methods of applying the biasing scores in the ASR system. Traditional ASR with a hybrid approach typically handles contextual biasing through shallow fusion of the biasing scores in a weighted finite state transducer (WFST), using methods such as class-based FST \cite{aleksic15_interspeech}, on-the-fly biasing with FST composition \cite{hall2015composition}, and HCLG and/or lattice boosting \cite{kocour2021boosting}, but moving to a so-called end-to-end (E2E) architecture \cite{li2019jasper} presents several new challenges. In this study, we focused on improving contextual biasing performance on these types of rare words by using trie-based biasing methods, enabling a system to favour their correct recognition and transcription in relevant contexts. 


\section{Related Work}
\label{sec:format}

Past research in E2E ASR for contextual biasing includes static knowledge incorporation with LM fusion \cite{toshniwal2018comparison, shan2019component}, dynamic knowledge incorporation with class-based weighted finite state transducer (WFST) \cite{he2019streaming, zhao2019shallow},  attention-based deep context \cite{pundak2018deep}, and trie-based deep biasing \cite{jain2020contextual, jung2022spell}. Applying trie-based biasing for conformer structure ASR systems is shown to be useful \cite{le2021contextualized}. Trie-based biasing shows success in a no-training-needed keyword boosting manner when using a CTC-conformer architecture \cite{jung2022spell}. Alternatively, deep biasing solutions that inject contextual information directly into the network \cite{le2021deep, jain2020contextual} are not reliant on context prefixes but do not work well when the biasing list gets larger; furthermore, this approach tends to underperform on rare words \cite{le2021deep, jain2020contextual}. Therefore, our work is based on a trie-based biasing mechanism.

One characteristic of business-related keywords is that they are often formed by an unconventional spelling or inclusion of non-alphanumeric characters; for example, they might be formed with alphanumeric sequences (e.g. “123OTwo”) or rare stylization (e.g. “l!sten\&\$ing”). Due to the rare occurrence of these targets in both acoustic model (AM) and language model (LM) training data, these low probability words may be pruned from the decoding beam, thus causing the trie-based biasing mechanism to be ineffective as they are not even showing up in the final beam, which means that the final biasing step will not be applied. Also, we notice that previous work \cite{le2021contextualized, jung2022spell, han2022improving} based on a synthetic keyword list with Librispeech might underestimate the difficulties of boosting these kinds of keywords since the synthetic keyword list in Librispeech does not include target examples with mixed alphanumeric characters.

\section{Contributions}
\label{sec:format}

Given that standard techniques in contextual biasing remain ineffective for terms with alphanumeric sequences and rare stylization, we present a two-step keyword boosting mechanism, as shown in Fig.\ref{fig:system_comparison}. Our approach first maps raw rare words into their normalized forms of unigrams and n-grams, then applies a biasing algorithm to these normalized targets rather than the raw single tokens. Finally, inverse text normalization is applied to map the normalized targets back to their original form before the model emits output transcripts. This method eliminates issues of missing hits when boosting raw targets. We also modified the boosting weight application logic to avoid over-boosting issues on n-grams. This improves our keyword recognition rate by 26\% relative on our proprietary in-domain dataset and 2\% relative on LibriSpeech.

\section{Data}
\label{sec:format}

\subsection{Internal test set and development set}
\label{ssec:subhead}
We collected an in-domain biasing target list from a subset of our proprietary customer-submitted target word dictionary. This in-house data set contains 1917 entries from our in-domain biasing target list. We also have metadata that includes the corresponding keyword type, such as company, person name, product name etc. We used the target words and their associated metadata to create contextually realistic sentences for which speech audio was generated using a Text-to-Speech (TTS) system. We compiled a test set of 1917 samples using this method.

We also compiled a development set following a similar approach but generated with a different TTS model. This development set is used for optimizing the boosting weight in the experiment.

\subsection{LibriSpeech test set}
\label{ssec:subhead}
We constructed the biasing lists by following the same methodology as \cite{han2022improving}, which effectively identified uncommon keywords that were unique to each book. The simulated biasing lists for test-clean and test-other are composed of 1171 and 1129 phrases, respectively. Since all the targets in the biasing list are unigrams and there are no alphanumeric/rare stylization tokens in LibriSpeech rare word list, this test set would not demonstrate the effect of boosting normalized n-gram targets. Therefore we took another step to simulate the real-world scenario: compound words in the rare word list are split by white space and treated as normalized n-gram targets for the biasing tasks.

\section{Experiment}
\label{sec:pagestyle}

\subsection{Baseline system}
\label{ssec:subhead}
The baseline system is an 18-layer Conformer-CTC model trained using Nvidia NeMo toolkit \cite{kuchaiev2019nemo} with 80-dimensional log-mel filterbank features as input and 1024 sub-word tokens as the output modeling unit using SentencePiece~\cite{kudo-2018-subword, kudo-richardson-2018-sentencepiece}. We use a beam search decoder with a 4-gram LM during inference using Pyctcdecode~\footnote{https://github.com/kensho-technologies/pyctcdecode/commit/\\a477d796e232b476ee8b877efba98aa2d822232e}~\cite{hannun2014first} without any boosting applied.

\begin{figure}[htb]
\caption{System comparison of default boosting and our n-gram boosting method}
\begin{minipage}[b]{1.0\linewidth}
  \centering
  \centerline{\includegraphics[width=8.5cm]{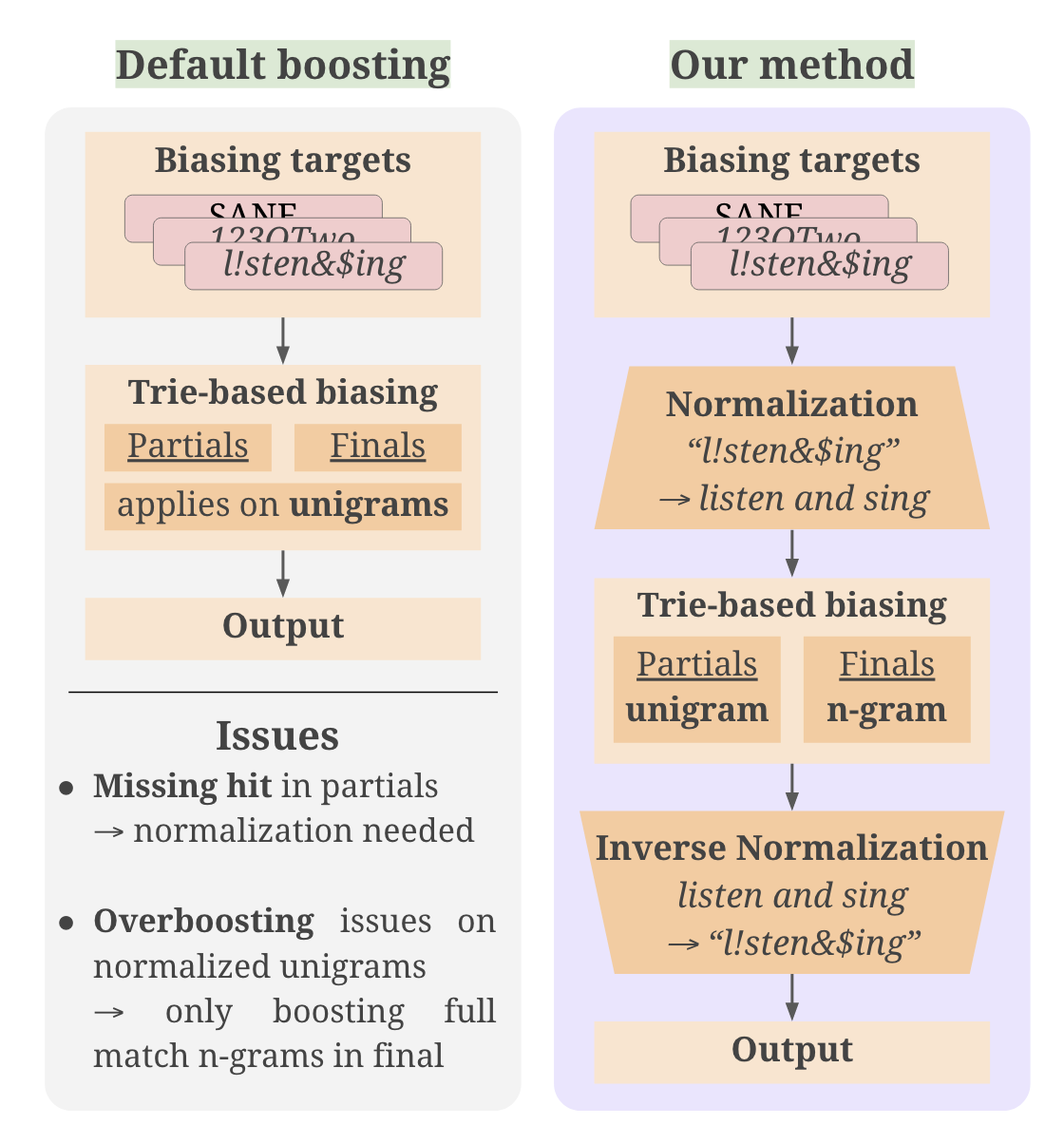}}
\end{minipage}
\label{fig:system_comparison}
\end{figure}

\subsection{Default boosting system}
\label{ssec:subhead}
The default boosting system is based on the same model mentioned in the baseline, but with the boosting mechanism applied on the given targets. We tested its boosting performance on raw targets and normalized targets in the two experiments respectively. This is a streaming model that emits partials in real-time, followed by a final hypothesis. When n-gram targets are given, the algorithm splits these n-grams into unigrams and applies boosting weights to each of these unigrams in partials and finals. To avoid over-boosting, boosting weight is only applied when the given unigram target has a log probability lower than -4.0 in our n-gram LM. Since targets are in their raw format, some of the targets might contain special characters and digits (e.g. ``l!sten\&\$ing'') which fall outside of our baseline model's tokenizer character list, thus the baseline model is incapable of correctly outputting these targets.

\subsection{N-gram boosting system}
\label{ssec:subhead}
This system uses the same baseline, but with the boosting mechanism applied on normalized targets, including a mix of unigrams and n-grams. The normalization module converts many different cases, including, but not limited to: digits (0-9, numerical sequence like 356) to their spoken form, symbols (\&, -, +, *, x) to their spoken form or eliminated, casing to lowercase form, compound words to n-grams separated by space, initialisms (e.g. IBM) to standalone letters, and acronyms (e.g. NASA) to lowercase form. These normalization steps are done by a semi-automatic approach where we leveraged the pronunciation dictionary from our internal hybrid ASR system, which has the appropriate pronunciation hints for each of these target words. We came up with the manually reviewed pronunciation data based on predictions made by Grapheme-to-phoneme (G2P) conversion models. Researchers can replicate this component with G2P system trained on standard entries in CMU Pronouncing Dictionary, which doesn’t need to be inferred from a hybrid ASR system. With the given pronunciation hints, we constructed the surface form representation of the target, followed by a manual review process. The same mapping is then saved and used in both the normalization and inverse-normalization processes before and after the trie-based boosting.

As illustrated in Fig.\ref{fig:system_comparison}, we first use our text normalization module to turn keywords into their normalized forms. Then we use trie-based biasing to add biasing scores to the beam search results. We applied the biasing scores to each unigram in partials to ensure individual tokens in the normalized target could end up in the final beam. In our default boosting system, n-grams are split up by space and treated as individual unigrams for boosting. However, we observed that this approach creates too many unigram targets when the biasing list has initialisms, which essentially results in boosting being applied on top of all single letters, leading to false positives in the results. This seems to be an issue that was not addressed in \cite{jung2022spell}. Therefore we adjust the boosting logic for n-gram boosting to only apply the biasing score to the full matches of the entire n-grams in the final output in order to avoid over-boosting issues. 

Finally, we apply inverse text normalization to map the boosted targets back to their expected written format. This is necessary as the target word in our use-case is the original format that the customer submitted for boosting, so we need to convert those boosted entries back to the expected written format.

\section{Results}
\label{sec:typestyle}
The trie-based biasing method does not require any training in advance, but with optimization on a development set it performs better \cite{jung2022spell}. We optimized the boosting weight for each target on a development set for in-domain results. We calculated word error rate (WER), U-WER, and B-WER \cite{le2021contextualized}, which refers to Unbiased-WER measured on words NOT IN the biasing list (U-WER) and Biased-WER measured on words IN the biasing list (B-WER) respectively. The goal is to improve on biased words without degrading unbiased words unduly. Biased terms for scoring only include the non-normalized, original terms (e.g. ``square1'' counts as B-WER, ``square'' and ``one'' count as U-WER).

\subsection{Over-boosting issues on normalized targets}

\begin{table}[]
\caption{Default biasing method easily causes over-boosting when targets involve initialisms or alphanumeric characters, resulting in higher WER \& U-WER, which motivates our adjustment for weight application in n-gram boosting
}
\resizebox{8cm}{!}{
\centering
\begin{tabular}{r|r|r|r}
\Xhline{3\arrayrulewidth}
System & WER & U-WER & B-WER\\ 
\Xhline{1\arrayrulewidth}
Baseline  & 17.43 & 16.81 & 21.56 \\ 
Default boosting & 20.29 (+2.86) & 21.39 (+3.48) & \textbf{12.92 (-8.64)} \\ 
N-gram boosting & \textbf{10.83 (-6.60)} & \textbf{10.44 (-6.37)} & 13.42 (-8.14)\\ 
\Xhline{3\arrayrulewidth}
\end{tabular}
}

\label{fig:exp1_results}
\end{table}

\begin{table}[]
\caption{False positives of standalone letters in default boosting with normalized target
}
\resizebox{8cm}{!}{
\begin{tabular}{c|c}
\Xhline{3\arrayrulewidth}
Reference   & they made a presentation about \textbf{AI analytics} \\ 
\Xhline{1\arrayrulewidth}
Baseline    & they made a presentation about \textbf{analytics} \\ 
Default     & they made a presentation about \textbf{ \textit{a i e e} analytics} \\ 
N-gram      & they made a presentation about \textbf{AI analytics} \\ 
\hline
\hline
Reference   & hi i am giving you a call back regarding \textbf{C3PO}\\ \hline
Baseline    & hi i am giving you a call back regarding \textbf{CT PO} \\ 
Default     & hi \textit{\textbf{i i i}} am giving you a call back regarding \textbf{C3PO} \\ 
N-gram      & hi i am giving you a call back regarding \textbf{C3PO} \\
\Xhline{3\arrayrulewidth}
\end{tabular}
}
\label{fig:exp1_examples}
\end{table}

\begin{figure}[htb]
\caption{System performance on our in-domain test set with different biasing approaches.)
}
  \centering
  \centerline{\includegraphics[width=8.5cm]{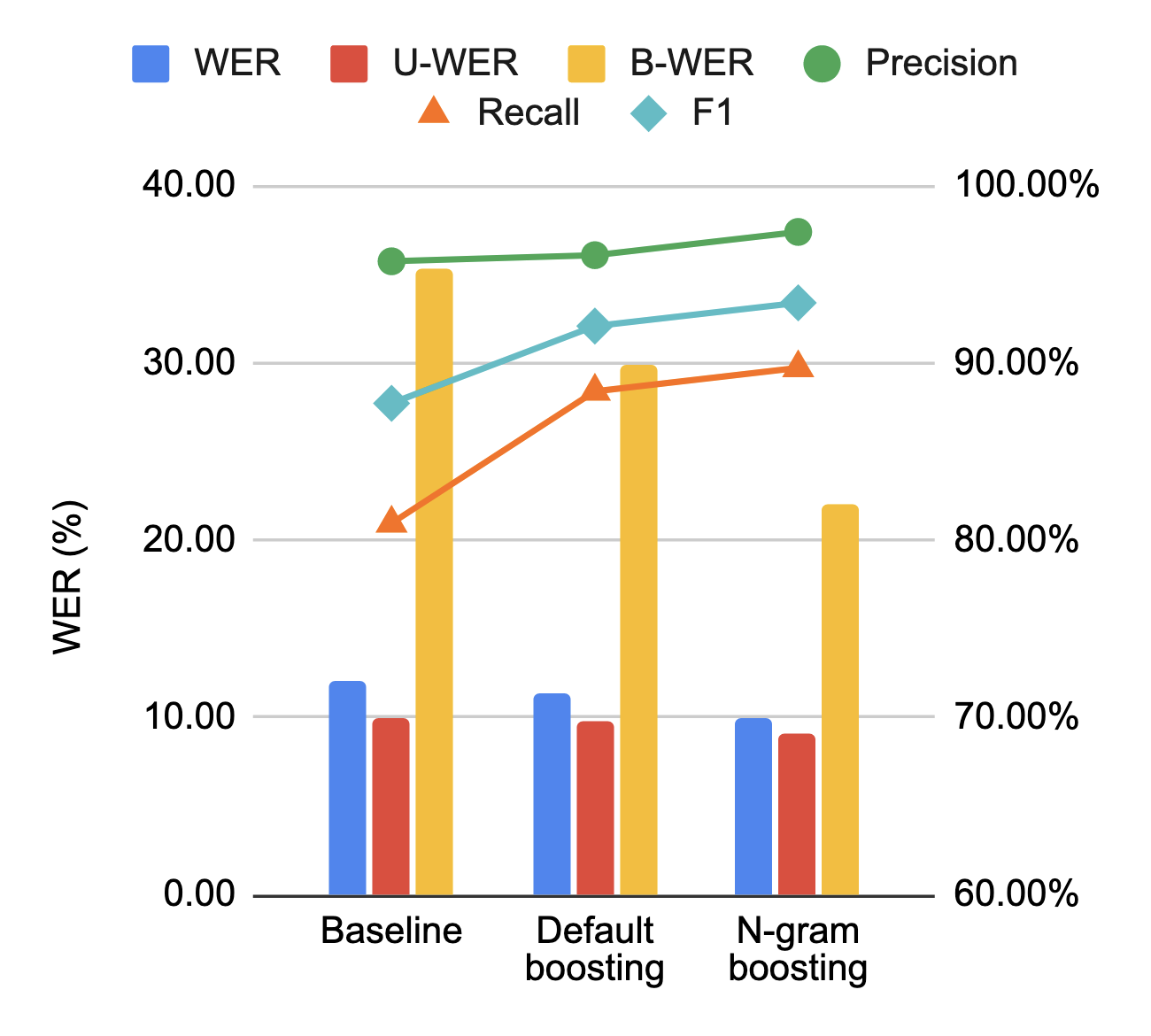}}
%
\label{fig:exp2_results}
\end{figure}

In the first experiment, we test whether there is any negative impact when we provide normalized targets for default boosting. We focused on a subset (365 samples) of the in-house test set, which contains targets with initialisms or alphanumeric characters. All the targets are normalized, and the boosting mechanism is applied to the normalized targets. As shown in Table \ref{fig:exp1_results}, the B-WER decreased to 12.92 when default boosting is applied. However, the default boosting approach also increases the U-WER (from 16.81 to 21.39) and overall WER (from 17.43 to 20.29) as a negative side effect. The reason for this degradation is that normalization turns most of these alphanumeric targets into multiple tokens, and since default boosting boosts these as individual tokens, the resulting boosting list consists of almost the entire list of individual standalone letters. The long list of biasing characters causes false positives of standalone letters to appear in transcripts, resulting in higher U-WER and WER, as illustrated in the examples in Table \ref{fig:exp2_results}. The corresponding WER statistics on the normalized targets are aligned with the WER statistics on the formatted targets.

We resolve this issue with n-gram boosting by adjusting the boosting logic to only boost full-match n-grams in finals in the n-gram boosting approach. In Table \ref{fig:exp1_results}, it is shown that with this modification, we are still able to get a similar amount of reduction in B-WER (-8.14\%) for all the target words while avoiding the negative side effect on U-WER. As a result, n-gram boosting reduces B-WER while not affecting U-WER, thus resulting in the overall improvement in WER (from 17.43 to 10.83).

\subsection{Missing hits in default boosting}

From the first experiment, we observed that default boosting on normalized targets suffers from the expanded biasing target list due to normalization. In the following experiments, we kept the biasing list for default boosting in its original format for comparison on the full in-house test set. Fig.\ref{fig:exp2_results} illustrates that n-gram boosting has the lowest B-WER among all three approaches, showing that normalization together with the adjusted boosting method is better than the default boosting. With further analysis of the decoding beam, it is found that some of the rare targets were not able to enter the final beam in partials/finals under default boosting, thus causing some missing hits when targets were not normalized. With the normalization, we turn these rare targets into relatively common tokens that can now benefit from the trie-based biasing technique as they are more likely to appear in the decoding beam. As a result, n-gram boosting reduces the B-WER from 29.96 to 22.12, which is 26\% relative reduction compared to that of the default boosting result.

\subsection{LibriSpeech coverage}
To assess the robustness of our approach on another domain without specific training, we also evaluated the boosting performance on the LibriSpeech test set. We decoded this test set by applying boosting on the simulated biasing list with a default boosting weight, which is picked through a grid search optimization process with our in-house development set. We observed 2\% relative B-WER improvements on top of the default boosting results, which suggests that our method can be broadly applied to out-of-domain targets and achieve improvements. Though the improvement on LibriSpeech is not as clear as we demonstrated in the in-house test set, we believe this is because LibriSpeech does not contain a variety of keyword-like targets that contains these rare stylizations or alphanumeric characters. 

Since the target word types that we focus on in our research, which are keywords in the business communication domain, do not exist in LibriSpeech, and thus are also not represented in any of the synthetic keyword lists used in \cite{le2021contextualized, jung2022spell, han2022improving}. The synthetic keyword list in these references contains words like “battery”, “leaping”, and “sixteenth” which are not the target words that our method is designed to improve upon. Here are the distribution statistics for comparison: tokens with symbols and/or alphanumeric substrings in our domain are 0.5\%, 0.4\%, respectively, while none of these exist in the LibriSpeech references. Mixed-case tokens account for 11.5\% of total words in our domain, but LibriSpeech doesn’t have mixed-cased tokens.

The lack of coverage of challenging tokens makes it more difficult to demonstrate the effectiveness of the normalization step. This suggests that a real-world biasing data set with a wide coverage of biasing targets is needed for a thorough assessment of biasing methods in similar research.

\subsection{Limitations}
The TTS samples used in our in-house dataset might not accurately reflect the acoustic characteristics of in-domain real-world data, even though we made efforts to simulate the semantic domain with the surrounding context according to entity type and the acoustic domain with various speaker characters such as gender, speaking rate, and pitch. Another limitation is that the normalization mapping requires a certain level of manual review; this could be a limiting factor for scaling in the future.

\section{Conclusions}
\label{sec:typestyle}
We observed that real-world targets for contextual biasing may often include n-grams with non-alphabetic characters or rare stylizations. These unusual orthographies have not been sufficiently addressed in previous research, and we identified that these targets present unique challenges when performing contextual biasing. However, by applying the n-gram boosting method we proposed, we improved B-WER by 26\% relative on our proprietary in-domain dataset and 2\% on LibriSpeech compared to the default boosting method. N-gram boosting is shown to be particularly useful on real-world keywords that involve alphanumeric characters and rare stylization.

\bibliographystyle{IEEEtran}
\bibliography{mybib}

\end{document}